\documentclass[11pt]{article}

\usepackage[preprint]{acl}

\usepackage{times}
\usepackage{latexsym}
\usepackage{algorithm}
\usepackage{multirow}
\usepackage{tabularx}
\usepackage{booktabs}
\usepackage{subcaption}
\usepackage{tabularx}
\usepackage{algpseudocode}
\usepackage{placeins}

\usepackage[T1]{fontenc}

\usepackage[utf8]{inputenc}

\usepackage{microtype}

\usepackage{graphicx}

\usepackage{amsmath}
\usepackage{amssymb}
\usepackage{amsthm}
\usepackage{bm}

\title{Lost in Interpolation: Why Predictive Feedback Fails in Diffusion Language Models}

\author{
  Lavanya Nigam$^{*}$, Ishaan Bansal$^{*}$, Aryan Sood$^{*}$, Vidit Aggarwal, Gaurav Kumar Nayak \\
  Indian Institute of Technology Roorkee, Roorkee, Uttarakhand, India \\
  \small
  \texttt{lavanya\_n@ma.iitr.ac.in}, 
  \texttt{ishaan\_b@ece.iitr.ac.in}, 
  \texttt{aryan\_s2@ee.iitr.ac.in}, \\
  \small
  \texttt{vidit\_a@mfs.iitr.ac.in},  
  \texttt{gauravkumar.nayak@mfs.iitr.ac.in}
}

\begin{document}
\maketitle

\begin{abstract}
Soft-masking accelerates the convergence of Masked Diffusion Language Models (MDLMs). Existing formulations build this blend with linear interpolation (LERP) in the raw embedding space, which implicitly treats that space as Euclidean. 
We analyze the embedding space of MDLMs and find that the mask and predicted-token embeddings maintain a near-constant angle of {$\approx$73\textdegree} throughout training, while embedding norms remain essentially flat across vocabulary-frequency rank.
These indicate a hyperspherical geometry, for which LERP is the wrong interpolation primitive. We introduce \emph{Spherical Soft-Masking (S-SM)}, a drop-in replacement that aggregates the top-$k$ predictions with a Fr\'{e}chet mean on the hypersphere and blends this mean with the mask direction using spherical linear interpolation (SLERP), then restores the native mask norm. We evaluate S-SM on continued pre-training of a released 169M-parameter MDLM checkpoint across a wide range of inference-time step budgets, SLERP feedback avoids the training degradation that LERP feedback induces and delivers MAUVE gains of up to 2× over the vanilla MDLM baseline and 27.5–56.1\% over TopK/LERP at various sampling budgets, alongside consistently lower generative perplexity (16.9–19.6\% over the baseline), while leaving output entropy and convergence essentially unchanged.
\end{abstract}

{
  \renewcommand{\thefootnote}{}
  \footnotetext{$^{*}$Equal contribution.}
}
\section{Introduction}

Masked diffusion language models (MDLMs) generate text by repeatedly denoising a fully masked sequence. At each denoising step, the backbone proposes a token for every still-masked position, and an unmasking rule decides which of these proposals to keep and which masked positions to leave untouched for the next step. This binary accept/reject rule is simple and trains well, but it throws away information the model knew before deciding whether to reveal the token or not. The \texttt{[MASK]} embedding left behind is identical at every position and every step and fails to carry the information gathered for predictions on prior steps.

\begin{figure}[t]
    \centering
    \includegraphics[width=\linewidth]{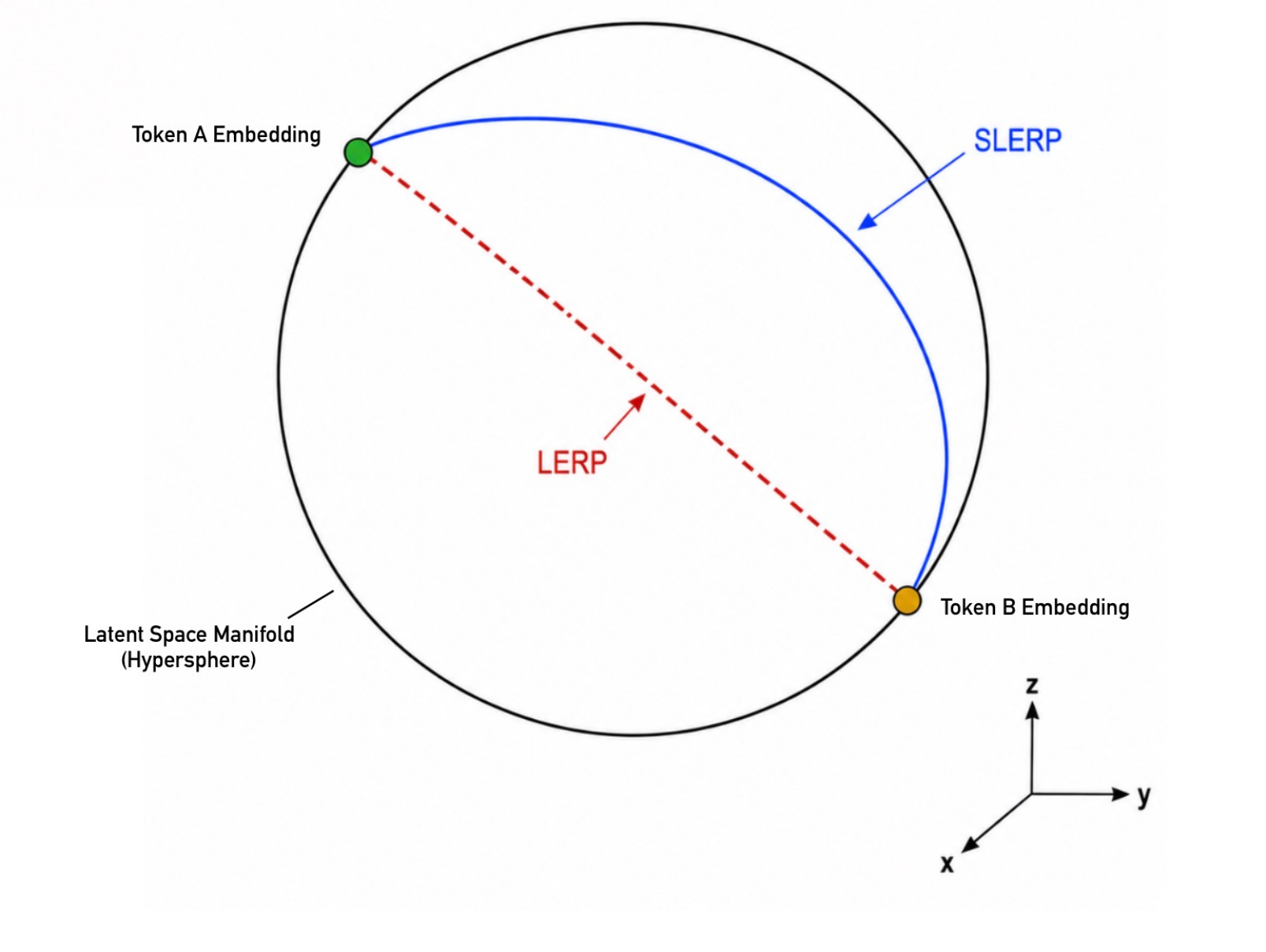}
    \caption{LERP and SLERP paths between two embeddings projected onto the unit hypersphere.}
    \label{fig:geometric_visualisation_slerp_and_lerp}
\end{figure}

Soft-masking (SM) was introduced to stop this information loss. SM replaces the binary unmask decision with a continuous confidence-weighted blend. For each retained masked position, the feedback embedding is constructed as a convex combination of the mask token embedding $\mathbf{m}$ and a weighted superposition of the top-$k$ predicted token embeddings. This single change has been shown to improve both perplexity and generation quality across model scales.

The current architectures build the blend as a linear combination of the mask embedding and the probability-weighted average of the top-k token embeddings. Linear interpolation is the natural choice if the underlying embedding space is Euclidean. Prior work on embedding geometry has repeatedly observed that trained token embeddings concentrate near a shell of roughly constant radius, i.e. a hypersphere, rather than filling the ambient Euclidean space. 
Interpolating linearly between two points on a hypersphere yields a point off the manifold, failing to accurately capture the semantics of the interpolation.
We empirically find that soft-masked MDLMs operate in a hyperspherical embedding space. The geodesic angle between the normalized mask embedding and the normalized mean of the model's top-k predictions holds steady at $\approx$73\textdegree across training, and the L2 norm of token embeddings is close to flat across GPT-2 frequency rank. Both observations point to a hyperspherical embedding geometry. 

We propose a more principled interpolation primitive for soft-masking in MDLMs. Spherical Soft-Masking (S-SM) replaces the Euclidean weighted mean of the top-k embeddings with a Fr\'{e}chet (Karcher) mean computed intrinsically on the unit hypersphere, and replaces the linear blend between the mask direction and this mean with spherical linear interpolation (SLERP) along the connecting geodesic. The result is rescaled to the mask token's native norm before being handed to the backbone. S-SM is a drop-in replacement for the linear feedback step and keeps the training objective, the unmasking rule and the confidence-based schedule identical

Our contributions are as follows:
\begin{enumerate}
    \item \textbf{Identifying Hyperspherical Embedding Geometry:} On the released MDLM checkpoint \cite{hersche2026soft}, we observe that the geodesic angle between the mask embedding and the top-$k$ mean holds near 73\textdegree{} and embedding norms are flat across frequency rank. This indicates a hyperspherical geometry, under which a linear blend of the embeddings lies off the spherical manifold.
    \item \textbf{S-SM, a geometric drop-in for linear feedback:} We  propose a replacement for linear feedback where we aggregate the top-$k$ embeddings with a Fr\'{e}chet mean on the sphere, blend via SLERP along the connecting geodesic, and rescale to the mask token's native norm.
    \item \textbf{Consistent gains across seeds and sampling budgets:} We show that S-SM nearly doubles MAUVE over the vanilla MDLM baseline at larger NFE budgets (256, 512) and improves MAUVE over TopK/LERP by over 50\% at the same budgets, while also avoiding the perplexity degradation LERP induces. On training with the equivalent learned confidence training scheme SLERP achieves higher MAUVE scores and lower generative perplexity across all sampling budgets and multiple seeds, while having nearly identical training perplexity. The learned confidence weight itself corroborates this: it converges to $\lambda\approx0.056$ under SLERP versus $\lambda\approx0.030$ under LERP (step 7k, $\approx$1.9$\times$), showing the optimizer trusts geometrically well-formed feedback more.
\end{enumerate}
\section{Related Work}
\label{sec:related_work}

\paragraph{Masked Diffusion Language Models.}
MDLMs frame text generation as iterative denoising over masked sequences, enabling parallel decoding as an alternative to autoregressive generation \citep{austin2021structured,dieleman2022continuous}. LLaDA \citep{nie2026large} scaled MDLMs to 7B parameters, while MaskGIT \citep{chang2022maskgit} validated parallel masked decoding for images. Because binary-decision MDLMs let early errors propagate irreversibly, ReMDM \citep{wang2026remasking} adds inference-time remasking, MDPO \citep{he2025mdpo} applies RL with confidence-based remasking, and DSFT \citep{chen2025dsft} extends MDLMs to mathematical reasoning via continued pre-training.

\paragraph{Soft Masking and Continuous Feedback.}
\citet{hersche2026soft} introduced Soft-Masking (SM), replacing binary unmasking with a confidence-weighted blend of mask and top-$k$ token embeddings, which serves as the direct precursor to our work. Related approaches include DMax \citep{chen2026dmax}, which applies linear interpolation to top-1 predictions under on-policy training, and context-aware initialization \citep{miao2025context}, which injects prompt priors to cut denoising steps. SSD-LM \citep{han2023ssd} showed continuous logit-simplex states carry semantic meaning, and \citet{jin2025role} validate soft-masking while identifying complementary failure modes, namely uniform corruption and token-wise marginal training, distinct from the embedding-space geometry issue we target here.

\paragraph{Riemannian and Hyperspherical Geometry.}
Several works treat transformer representations as living on a hypersphere. nGPT \citep{loshchilov2025ngpt} normalizes all embeddings and states to unit norm and interprets residual updates as SLERP rotations. anTransformer \citep{franke2026learning} achieves similar speedups via approximate normalization. Harmonizing Geometry and Uncertainty \citep{dosi2025harmonizing} and Discrete Stochastic Localization \citep{wu2026discrete} apply von Mises, Fisher noise and hypersphere mappings respectively, sharing our geometric motivation.

\paragraph{Continuous Diffusion on Manifolds.}
RDLM \citep{jo2026continuous} formally connects masked diffusion to geodesic flow on $\mathbb{S}^{d-1}$, directly motivating our LERP-to-SLERP substitution. Fisher Flow Matching \citep{davis2024fisher} and GIDD \citep{von2025generalized} similarly reparameterize discrete categories as hyperspherical flows. S-FLM \citep{deschenaux2026sflm} applies SLERP-based angular noise during pretraining, which forms the closest architectural parallel to our continued pre-training approach.

\section{Background}

\paragraph{Vocabulary and embeddings.} Let $V$ be the vocabulary size and $D$ the embedding dimension. The backbone's token-embedding table is $E \in \mathbb{R}^{V \times D}$, where row $E[v] \in \mathbb{R}^D$ the embedding of token $v$. We write $\mathbf{m} = E[\texttt{[MASK]}]$ for the dedicated mask embedding and $r_m = \lVert \mathbf{m} \rVert$ for its norm. For any vector $\mathbf{v}$ we write $\hat{\mathbf{v}} = \mathbf{v}/\lVert \mathbf{v} \rVert$ for its projection onto the unit sphere $\mathbb{S}^{D-1}$.

\paragraph{Forward and reverse process.} A clean sequence $\mathbf{x}_0 \in \{1,\dots,V\}^L$ of length $L$ is corrupted into $\mathbf{x}_t$ by independently replacing each token with $\texttt{[MASK]}$ with probability $1-\alpha_t$, where $\alpha_t$ decreases monotonically from $1$ at $t=0$ to $0$ at $t=T$. We use the standard linear schedule $\alpha_t = 1 - t/T$ throughout this work, for both pretraining and continued pre-training. The backbone $g_\theta : \mathbb{R}^{L\times D} \to \mathbb{R}^{L \times V}$ maps the (possibly soft-masked) embedded sequence to per-position token distributions $\mathbf{p}^{1:L}_{t-1} = g_\theta(\mathbf{x}_t)$, and denoising proceeds by iteratively sampling and partially unmasking positions until $t=0$. The number of denoising steps used at inference/generation time, $T$, need not equal the number of training corruption levels. We report it relative to sequence length as an NFE (Number of Function Evaluations) budget $T/L$, with $T=L$ corresponding to the full, unconstrained budget of on average one token revealed per step.

\paragraph{Soft-masking (SM).} Rather than leaving a retained mask position as the bare embedding $\mathbf{m}$, SM overwrites it with a convex combination of $\mathbf{m}$ and the top-$k$ predicted token embeddings from the previous step, weighted by their renormalized softmax probabilities $\pi_i$:
$$
\tilde{\mathbf{x}}^{l}_{t-1} \;=\; \texttt{FEEDBACK}\big(\mathbf{m}, \{(\mathbf{v}_i,\pi_i)\}_{i\in\text{top-}k}, \lambda\big),
$$
applied only at positions where the model chose to keep the mask. Here $\mathbf{v}_i = E[i]$ is the embedding of the $i$-th top-$k$ candidate and $\lambda \in [0,1)$ is a confidence-based mixing weight derived from the predictive entropy $H(\mathbf{p}^l_{t-1})$ of the position (the interpolation operators $\mathcal{F}_{\text{LERP}}, \mathcal{F}^\ast$ are themselves defined over the closed interval $[0,1]$; the schedule below always yields values strictly less than 1):
\begin{equation}
\label{eq:lambda_confidence}
\lambda(\mathbf{p}_{t-1}) = \omega_s \cdot \sigma\!\big(\omega_a\,(-H(\mathbf{p}_{t-1}) - \omega_b)\big),
\end{equation}
with three learnable scalars $(\omega_s,\omega_a,\omega_b)$ shared across positions and trained jointly with the backbone. Intuitively: high confidence (low entropy) pushes $\lambda$ toward $\omega_s$, injecting more of the model's own prediction; low confidence keeps $\lambda$ near $0$, i.e. close to the plain mask. We refer to the family of instantiations of $\texttt{FEEDBACK}$ as soft-masking feedback operators, and to the specific choice of how the top-$k$ embeddings are aggregated and blended with $\mathbf{m}$ as the interpolation geometry.

\paragraph{The linear (LERP) operator.} The existing formulation aggregates the top embeddings $k$ with a Euclidean weighted mean $\boldsymbol\mu_{\text{LERP}} = \sum_{i} \pi_i \mathbf{v}_i$ and blends it with the mask embedding by linear interpolation:
$$
\mathcal{F}_{\text{LERP}}(\mathbf{m}, \{(\mathbf{v}_i,\pi_i)\}, \lambda) \;=\; (1-\lambda)\,\mathbf{m} + \lambda\,\boldsymbol\mu_{\text{LERP}}.
$$
We refer to this as the TopK/LERP baseline. It is well-defined and differentiable, and it recovers the two sensible extremes: $\lambda=0$ gives back the plain mask (vanilla MDLM), and $\lambda\to1$ gives back the (Euclidean) aggregate of the model's own top-$k$ prediction.

\begin{figure}[t]
    \centering
    \includegraphics[width=\linewidth]{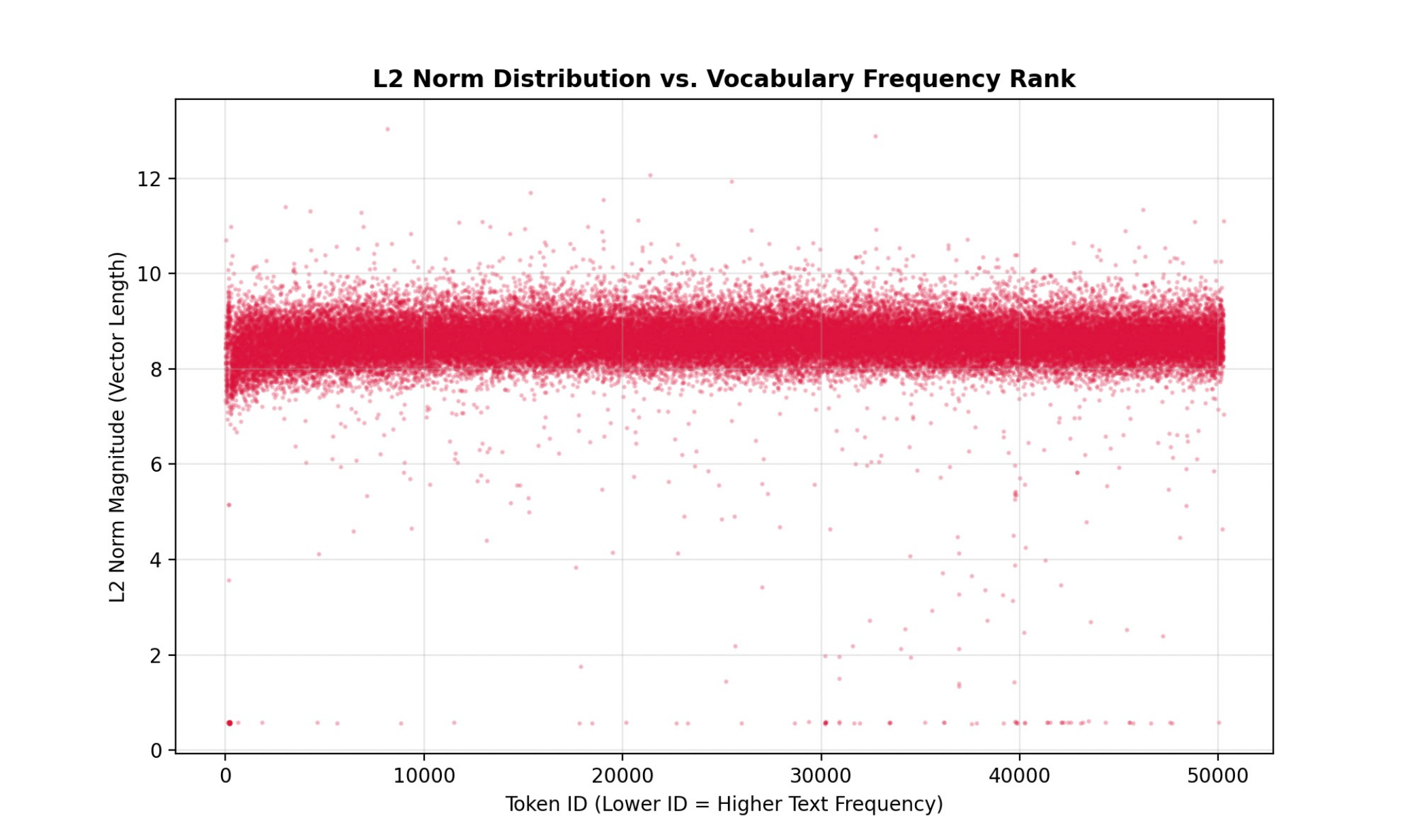}
    \caption{$L_2$ norm of every token embedding $E[v]$ against its GPT-2 byte-pair-encoding frequency rank on OpenWebText.}
    \label{fig:L2_norm_of_token_embeddings}
\end{figure}

\section{Problem Formulation}

\paragraph{Norm vs. frequency.} Figure \ref{fig:L2_norm_of_token_embeddings} show token embedding lie on a hyperspherical shell. Norms are essentially flat across four orders of magnitude of frequency rank, with no systematic trend for common versus rare tokens. 

\paragraph{Geodesic angle diagnostic.} For every masked position visited during training, we compute the geodesic angle $\theta = \arccos\langle \hat{\mathbf{m}}, \hat{\boldsymbol\mu} \rangle$ between the normalized mask embedding and the normalized Euclidean mean of the top-$k$ predictions. Averaged across positions and steps, $\theta$ holds at $\approx$73\textdegree ($\approx 1.27$ rad) essentially for the entire duration of training.

Both diagnostics point the same way: embeddings live on (or very near) a shell of near-constant radius, and the mask direction sits at a substantial, stable angle from where the model's own predictions point. This is precisely the geometry under which a straight-line blend is the wrong primitive.

\subsection{Design targets for a geometry-respecting operator}

We want a replacement operator $\mathcal{F}^\ast$ satisfying three properties:

\begin{itemize}
\item[(P1)] \textbf{Endpoint fidelity:} $\mathcal{F}^\ast(\mathbf{m},\cdot,0)=\mathbf{m}$ and $\mathcal{F}^\ast(\mathbf{m},\cdot,1) = r_m\,\hat{\mathbf{v}}_{\text{agg}}$ for some $\hat{\mathbf{v}}_{\text{agg}}\in\mathbb{S}^{D-1}$.
\item[(P2)] \textbf{Norm compatibility:} $\lVert\mathcal{F}^\ast\rVert = r_m$ for every $\lambda\in[0,1)$, matching the norm range the backbone was pretrained on.
\item[(P3)] \textbf{Geodesic interpolation:} the output direction traces the great-circle arc on $\mathbb{S}^{D-1}$ between the mask direction and the aggregate target, rather than a chord.
\end{itemize}
\begin{figure*}[t]
    \centering
    \includegraphics[width=0.95\textwidth]{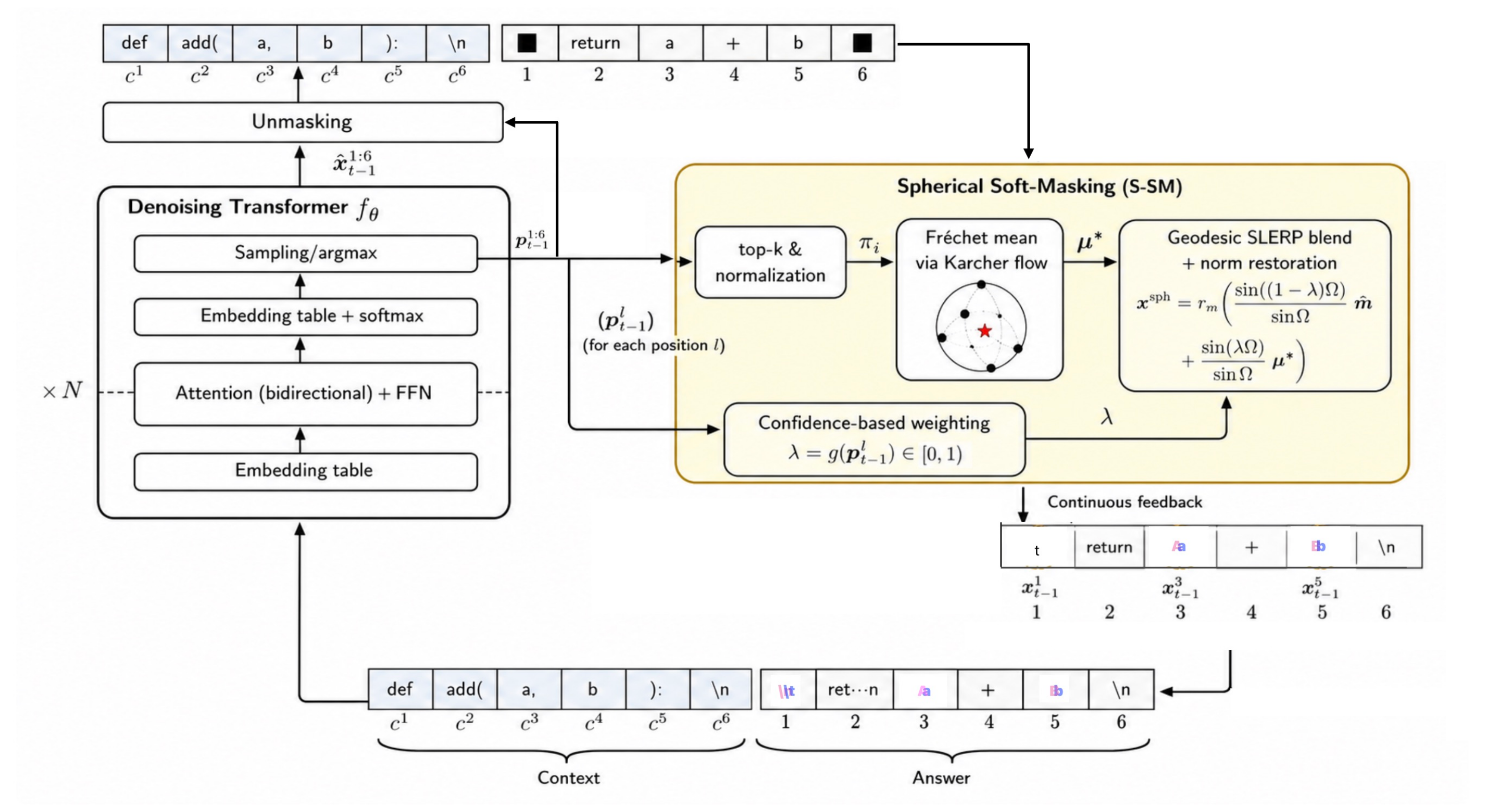}
    \caption{Iterative denoising in MDLMs using spherical soft-masking (S-SM).}
    \label{fig:slerp_architecture_diagram}
\end{figure*}

Simply renormalizing $\mathcal{F}_{\text{LERP}}$'s output to norm $r_m$ satisfies (P2) but not (P3): the direction still traces a chord-projected curve, not the geodesic, and the construction is numerically unstable whenever $\mathbf{m}$ and $\boldsymbol\mu_{\text{LERP}}$ are close to antipodal. Section~\ref{sec:method} constructs an operator satisfying all three.

\section{Methodology}
\label{sec:method}

S-SM replaces $\mathcal{F}_{\text{LERP}}$ with a two-stage operator that (1) aggregates the top-$k$ candidates \emph{on the sphere} rather than in raw Euclidean space, and (2) blends the result with the mask direction along the sphere's surface rather than through a straight line, before restoring the mask token's native norm. Intuitively, instead of averaging points as if they lived in flat space and cutting a straight line between them, S-SM treats every embedding as a direction on a globe and moves along the great-circle arc between two directions, the spherical equivalent of a straight line, which never leaves the surface. Figure~\ref{fig:slerp_architecture_diagram} illustrates the resulting denoising loop. The full geometric construction (Riemannian log/exp maps, Karcher iteration, closed-form SLERP) is given in Appendix~\ref{app:construction}.

\subsection{Geometry on \texorpdfstring{$\mathbb{S}^{D-1}$}{S\^{}(D-1)}}
We project token embeddings to the hypersphere via $L_2$ normalization, $\hat{\mathbf{v}} = \mathbf{v} / \lVert\mathbf{v}\rVert$, and evaluate proximity using the geodesic distance:
\begin{equation}
d_g(\mathbf{a}, \mathbf{b}) = \arccos \langle\mathbf{a}, \mathbf{b}\rangle.
\label{eq:geodesic}
\end{equation}
For any base point $\bm{\mu} \in \mathbb{S}^{D-1}$, the logarithmic and exponential maps are defined as:
\begin{equation}
\log_{\bm{\mu}}(\mathbf{v}) = \frac{\theta}{\sin\theta} \big(\mathbf{v} - \cos\theta \, \bm{\mu}\big), \quad \theta = \arccos \langle\bm{\mu}, \mathbf{v}\rangle,
\label{eq:logmap}
\end{equation}
\begin{equation}
\exp_{\bm{\mu}}(\mathbf{u}) = \cos(\lVert\mathbf{u}\rVert)\,\bm{\mu} + \sin(\lVert\mathbf{u}\rVert) \frac{\mathbf{u}}{\lVert\mathbf{u}\rVert}, \quad \mathbf{u} \in T_{\bm{\mu}}\mathbb{S}^{D-1}.
\label{eq:expmap}
\end{equation}
The $\log$ map projects spherical coordinates into the flat tangent space $T_{\bm\mu}\mathbb{S}^{D-1}$; the $\exp$ map retracts vectors back onto the sphere along a local geodesic path. Both maps are used only as internal building blocks of the feedback computation.

\subsection{Spherical aggregation of top-$k$ predictions}
\label{sec:frechet-mean}

Rather than the Euclidean weighted mean $\boldsymbol\mu_{\text{LERP}}=\sum_i \pi_i \mathbf{v}_i$, we compute a \emph{Fr\'{e}chet mean}: the direction $\boldsymbol\mu^\star$ on the unit sphere that minimizes the (squared, confidence-weighted) angular distance to each of the top-$k$ candidate directions $\hat{\mathbf{v}}_i$:
\begin{equation}
\boldsymbol\mu^{\star} = \arg\min_{\bm{\mu} \in \mathbb{S}^{D-1}} \sum_{i \in \mathrm{top\text{-}}k(\mathbf{p})} \pi_i \, d_g(\bm{\mu}, \hat{\mathbf{v}}_i)^2.
\label{eq:frechet}
\end{equation}
Equation~\eqref{eq:frechet} has no closed-form solution, so we approximate it with $N_{\text{iter}}=3$ steps of an iterative procedure (Karcher flow) that starts at the top-1 prediction and repeatedly nudges toward the confidence-weighted consensus direction:
\begin{equation}
\boldsymbol\mu^\star \leftarrow \exp_{\boldsymbol\mu^\star} \left( \sum_{i \in \mathrm{top\text{-}}k(\mathbf{p})} \pi_i \log_{\boldsymbol\mu^\star}(\hat{\mathbf{v}}_i) \right).
\label{eq:karcher}
\end{equation}
Because the softmax weights $\pi_i$ are typically peaked on the top-1 candidate, this converges in very few iterations (Appendix~\ref{app:construction}).

\subsection{Spherical blend and norm restoration}
\label{sec:slerp-blend}

Given the mask direction $\hat{\mathbf{m}}$ and the aggregated target $\boldsymbol\mu^\star$, we interpolate between them using \textbf{spherical linear interpolation (SLERP)}: instead of a straight-line blend $(1-\lambda)\hat{\mathbf{m}}+\lambda\boldsymbol\mu^\star$, which cuts through the interior of the sphere and shrinks in norm, SLERP produces a unit vector at every $\lambda\in[0,1]$ by construction. Setting $\Omega = \arccos\langle\hat{\mathbf{m}}, \boldsymbol\mu^\star\rangle$, we compute:
\begin{equation}
\mathbf{s} = \frac{\sin((1-\lambda)\Omega)}{\sin\Omega} \hat{\mathbf{m}} + \frac{\sin(\lambda\Omega)}{\sin\Omega} \boldsymbol\mu^{\star}.
\label{eq:slerp}
\end{equation}
We then rescale this unit vector to the mask token's native norm $r_m = \lVert \mathbf{m} \rVert$ before handing it to the backbone:
\begin{equation}
\mathbf{x}^{\text{sph}} = r_m\,\mathbf{s}.
\label{eq:rescale}
\end{equation}
The result is a feedback vector that (i) equals the plain mask at $\lambda=0$, (ii) equals the aggregated prediction (at native norm) as $\lambda\to1$, and (iii) has \emph{exactly} $r_m$ for every intermediate $\lambda$.

\paragraph{Numerical safeguards.} All inner products used inside $\arccos$ are clamped to $[-1+\epsilon, 1-\epsilon]$, and the denominator $\sin\Omega$ in Equation~\eqref{eq:slerp} is clamped to $[\epsilon, \infty)$. When the mask and target directions are nearly identical ($\Omega < \delta$), SLERP becomes numerically ill-conditioned; we detect this case and fall back to a normalized linear blend, $\mathbf{s} = \big((1-\lambda)\hat{\mathbf{m}} + \lambda\boldsymbol\mu^\star\big) / \lVert (1-\lambda)\hat{\mathbf{m}} + \lambda\boldsymbol\mu^\star \rVert$, which is safe precisely because the two directions are already almost the same.

\paragraph{Drop-in property.} S-SM changes only the interior of $\texttt{FEEDBACK}$: the confidence schedule, unmasking rule, training objective, and two-pass training procedure are all unchanged (Section~\ref{sec:training-procedure}). Every operation is differentiable, so gradients flow through S-SM exactly as they do through $\mathcal{F}_{\text{LERP}}$.

\section{Experiments}

We evaluate S-SM against two baselines, the no-feedback \textbf{vanilla MDLM} ($\lambda\equiv0$) and the \textbf{TopK/LERP} operator, under an identical training recipe, varying only the interpolation geometry inside $\texttt{FEEDBACK}$. All experiments use the 169M-parameter Diffusion Transformer (DiT) backbone and tokenizer of the underlying MDLM (GPT-2 BPE, GPT-2 BPE, vocabulary 50,257, extended with a dedicated [MASK] token for a total vocabulary of 50,258), pre-trained and subsequently continued-pretrained on OpenWebText with sequence length $L=1024$, $k=3$ top-$k$ candidates, and $N_{\text{iter}}=3$ Karcher steps unless stated otherwise.

\begin{figure}[t]
    \centering
    
        \centering
        \includegraphics[width=\linewidth]{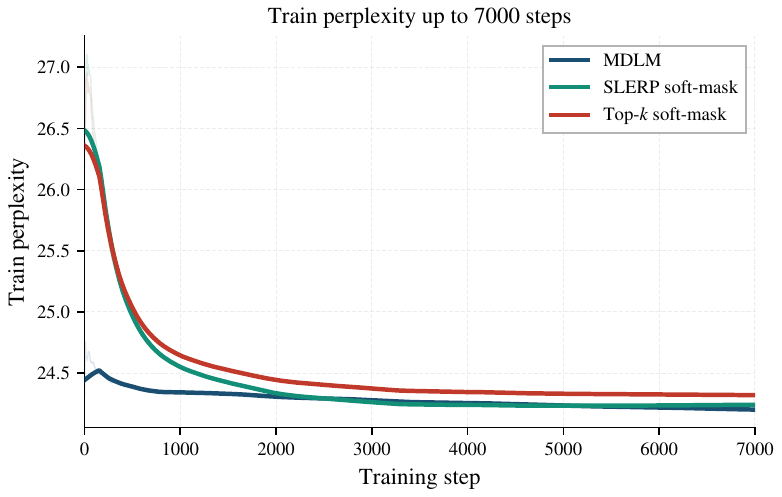}
        \caption{Training perplexity for the vanilla MDLM control, TopK/LERP, and SLERP feedback, up to step 7000.}
        \label{fig:train-ppl}
    
    \hfill
    
\end{figure}

\subsection{Training procedure}
\label{sec:training-procedure}

\textbf{Two-pass training.} As is standard for soft-masking, one training step consists of a \emph{gradient-free} forward pass that produces the previous step's predictive distribution $\tilde{\mathbf{p}}^{1:L}_{t-1}$, followed by construction of the soft-masked input via $\texttt{FEEDBACK}$, followed by a \emph{second} forward pass with gradients through which the backbone and the confidence parameters $(\omega_s,\omega_a,\omega_b)$ are jointly updated. This recipe is shared, unchanged, by every feedback operator we test.

\textbf{Stochastic time band (shared across all configurations).} Following the narrower-timestep-sampling recommendation of prior block-diffusion work, we draw the noise level $t$ for every soft-masking training step from a bounded uniform distribution $t\sim\mathcal{U}(b_l,b_h)$ with $(b_l,b_h)=(0.2,0.8)$, rather than the full $[0,1]$ range used for the underlying (non-soft-masked) diffusion loss. This \emph{time band} is applied identically to the vanilla MDLM control (where it is a no-op on the feedback, since $\lambda\equiv0$), to TopK/LERP, and to S-SM; the only source of difference between conditions is the interpolation geometry itself. Soft-masking is additionally activated stochastically with probability $p_{sm}=0.5$, so that the backbone also sees plain (non-soft-masked) inputs during training and does not collapse onto always expecting feedback.

\subsection{Continued Pre-Training}

We start from a publicly released 169M-parameter MDLM checkpoint pretrained on OpenWebText for 1M steps and continue training for a fixed \textbf{7,000-step} budget under each feedback configuration, so that all comparisons isolate the effect of the interpolation geometry rather than total training time. We test a \textbf{learned} mixing weight (via the confidence schedule of Equation~\eqref{eq:lambda_confidence}, initialized near $\lambda\approx0.13$-$0.14$). Full optimizer, learning-rate, and compute details are given in Appendix.

All generation is scored with a GPT-2-large evaluator. Results at $T\in\{64,128,256\}$ are reported as mean $\pm$ sample standard deviation over 3 seeds; $T{=}512$ reflects a single seed. Per-seed values and 95\% confidence intervals for the 3-seed budgets are given in Appendix~\ref{app:seed_sweep}.

\begin{table*}[t]
\centering
\small
\begin{tabular}{lcccccc}
\toprule
& \multicolumn{2}{c}{\textbf{MDLM (baseline)}} & \multicolumn{2}{c}{\textbf{TopK/LERP}} & \multicolumn{2}{c}{\textbf{SLERP (ours)}} \\
\cmidrule(lr){2-3} \cmidrule(lr){4-5} \cmidrule(lr){6-7}
\textbf{NFE (T)} & Gen PPL $\downarrow$ & MAUVE $\uparrow$ & Gen PPL $\downarrow$ & MAUVE $\uparrow$ & Gen PPL $\downarrow$ & MAUVE $\uparrow$ \\
\midrule
1/16 (64)  & $87.25\pm0.82$ & $0.0086\pm0.0002$ & $82.58\pm0.45$ & $0.0091\pm0.0007$ & $\mathbf{72.50\pm0.62}$ & $\mathbf{0.0117\pm0.0008}$ \\
1/8 (128)  & $73.15\pm0.41$ & $0.0116\pm0.0003$ & $69.07\pm0.10$ & $0.0148\pm0.0001$ & $\mathbf{59.76\pm0.34}$ & $\mathbf{0.0189\pm0.0020}$ \\
1/4 (256)  & $66.18\pm0.08$ & $0.0176\pm0.0031$ & $62.46\pm0.35$ & $0.0231\pm0.0021$ & $\mathbf{53.73\pm0.53}$ & $\mathbf{0.0354\pm0.0055}$ \\
1/2 (512)$^\dagger$  & 63.43 & 0.0199 & 59.46 & 0.0264 & \textbf{51.01} & \textbf{0.0412} \\
\bottomrule
\end{tabular}
\caption{Unconstrained generation quality vs.\ NFE budget (learned $\lambda$, 7k continued pre-training steps). Rows for $T\in\{64,128,256\}$ report mean $\pm$ sample standard deviation across 3 seeds; per-seed values and 95\% CIs are given in Appendix~\ref{app:seed_sweep}. $^\dagger T{=}512$ reflects a single seed only.}
\label{tab:gen-quality-nfe}

\end{table*}

SLERP outperforms both TopK/LERP and the vanilla MDLM baseline on Gen PPL and MAUVE at every budget tested, and the ranking is consistent across all 3 seeds at $T\in\{64,128,256\}$ (Appendix~\ref{app:seed_sweep}). SLERP improves mean MAUVE over the MDLM baseline by 36.0–107.0\% and over TopK/LERP by 27.7–56.1\%, with the largest gains at T=256 and T=512; it also improves mean Gen PPL by 16.9–19.6\% and 12.2–14.2\% respectively across all four NFE budgets. Entropy at T=64 (mean over 3 seeds): MDLM 5.5849, TopK/LERP 5.5837, SLERP 5.5658, comparable across configurations, indicating SLERP's quality gains are not a diversity trade-off, as detailed in Appendix~\ref{app:seed_sweep} (Table~\ref{tab:app_seed_ci}).

\section{Ablations}
\label{sec:ablations}

\begin{figure}[t]
    \centering
    \includegraphics[width=\linewidth]{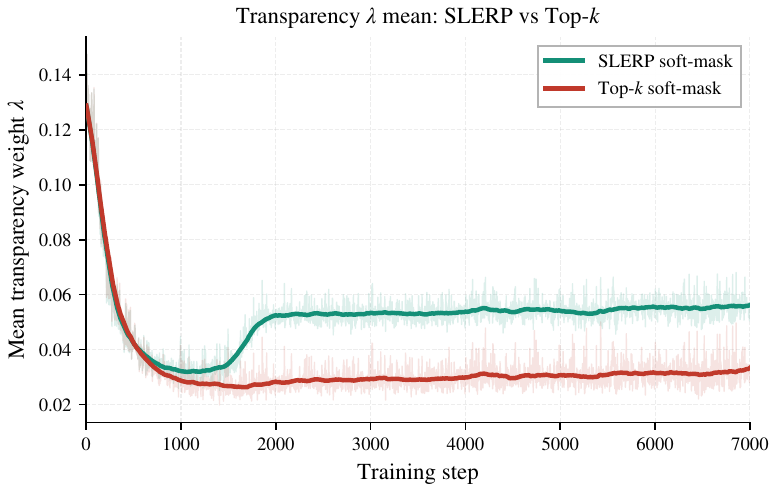}
    \caption{Mean learned $\lambda$ vs.\ training step (SLERP vs. TopK/LERP), up to step 7000.}
    \label{fig:learned_lambda_feedback}
\end{figure}

\begin{table}[t]
\centering
\small
\begin{tabular}{lcccc}
\toprule
\textbf{Feedback} & \textbf{PPL} & \textbf{NLL} & $\lambda_{\text{mean}}$ & \textbf{SLERP $\angle$} \\
\midrule
Vanilla ($\lambda=0$) & 24.20 & 3.187 & 0.000 & --- \\
LERP                  & 24.32 & 3.191 & 0.030 & --- \\
SLERP                 & 24.24 & 3.188 & 0.056 & $73.1^{\circ}$ \\
\bottomrule
\end{tabular}
\caption{Feedback-type comparison under learned $\lambda$ at step 7000 (averaged over 3 seeds).}
\label{tab:feedback_type_comparison}
\end{table}

\subsection{Karcher Iterations}
\label{sec:ablation-karcher}

Holding $k=3$ fixed, we vary the number of Karcher fixed-point iterations used to compute the Fr\'echet mean (Table~\ref{tab:karcher_topk_ablations}, left). At the matched checkpoint (step 1181), the three settings differ by at most 0.11 PPL, with no monotonic trend across $n_{\text{iter}}\in\{1,3,5\}$. This is consistent with the peaked top-$k$ weights argument of \citet{hersche2026soft}: the softmax over top-$k$ logits concentrates mass on the top-1 token, so the Karcher flow is already near its fixed point after one iteration. While $n_{\text{iter}}=5$ attains the lowest PPL in this ablation (24.47) and $n_{\text{iter}}=1$ the cheapest, the gap between all three settings is within noise. We adopt $n_{\text{iter}}=3$ as the best accuracy-compute balance: it captures the same near-fixed-point accuracy as $n_{\text{iter}}=5$ while requiring three-fifths the trigonometric-operation cost per masked position, and provides a small safety margin over $n_{\text{iter}}=1$ against configurations where the top-$k$ distribution is less peaked and convergence is slower.

\subsection{Top-\emph{k} Feedback Width}
\label{sec:ablation-topk}

With $n_{\text{iter}}=3$ fixed, we vary $k$ (Table~\ref{tab:karcher_topk_ablations}, right). The pattern is clearer: $k=1$ trails $k=3$ by 0.17 PPL, while $k=3$ and $k=5$ stay within 0.12 PPL of each other. The largest marginal gain is from $k=1\rightarrow3$, diminishing sharply from $k=3\rightarrow5$. This mirrors the pattern reported by \citet{hersche2026soft} for their own top-$k$ ablation. Since Fr\'echet-mean computation scales linearly in $k$, and wider superposition has been reported elsewhere to degrade downstream performance, $k=3$ captures nearly all of the accuracy gain available from top-$k$ aggregation at less compute than $k=5$. We therefore adopt $k=3$, paired with $n_{\text{iter}}=3$, as the accuracy-compute operating point used throughout this work.

\begin{table}[t]
\centering
\small
\setlength{\tabcolsep}{4pt}
\begin{minipage}{0.48\columnwidth}
\centering
\begin{tabular}{cccc}
\toprule
\textbf{$n_{\text{iter}}$} & \textbf{PPL} & \textbf{NLL} & $\lambda_{\text{mean}}$ \\
\midrule
1 & 24.48 & 3.198 & 0.059 \\
3 & 24.58 & 3.202 & 0.035 \\
5 & 24.47 & 3.197 & 0.060 \\
\bottomrule
\end{tabular}
\end{minipage}%
\hfill
\begin{minipage}{0.48\columnwidth}
\centering
\begin{tabular}{cccc}
\toprule
\textbf{$k$} & \textbf{PPL} & \textbf{NLL} & $\lambda_{\text{mean}}$ \\
\midrule
1 & 24.75 & 3.209 & 0.030 \\
3 & 24.58 & 3.202 & 0.035 \\
5 & 24.46 & 3.197 & 0.034 \\
\bottomrule
\end{tabular}
\end{minipage}
\caption{Karcher-iteration (left) and top-$k$ width (right) ablations at the common checkpoint, step 1181.}
\label{tab:karcher_topk_ablations}
\end{table}

The $k=3$, $n_{\text{iter}}=3$ reference run continued training to step 7000. It ultimately reached PPL 24.24, NLL 3.188, and $\lambda_{\text{mean}}$ 0.056 (Table~\ref{tab:feedback_type_comparison}).

\subsection{Feedback Regime and Embedding Geometry}
\label{sec:ablation-feedback-regime}

We compare the $\lambda=0$ vanilla control against learned-$\lambda$ LERP and learned-$\lambda$ SLERP. This comparison uses centre $-4.0$ and step $\approx7000$ where applicable. All three converge at comparable final perplexity, indicating that under a learnable $\lambda$ the optimizer suppresses both feedback mechanisms toward the no-feedback regime. Nevertheless, SLERP's transparency head sustains roughly twice LERP's feedback magnitude for the same final loss, and its Fr\'echet-mean target remains well clear of the near-collinear regime in which SLERP would degenerate to a Euclidean blend. Table~\ref{tab:feedback_type_comparison} reports the exact $\lambda_{\text{mean}}$ and angle values. Figure \ref{fig:learned_lambda_feedback} shows this divergence emerging over the course of training and stabilizing well before step 7000.

In Figure \ref{fig:train-ppl}, SLERP's training-perplexity curve sits below both the vanilla control and TopK/LERP for most of the training window, while TopK/LERP's curve remains elevated relative to the no-feedback baseline, consistent with the fact that linear interpolation is geometrically wrong, and embeddings lie on a hypersphere.

\subsection{Seed Robustness}

\label{sec:seed_robustness}

To confirm the ranking in Table~\ref{tab:gen-quality-nfe} is not an artifact of a single training/generation seed, we repeat generation under 3 seeds for all three configurations (MDLM, TopK/LERP, SLERP) at $T\in\{64,128,256\}$; these per-seed runs are what Table~\ref{tab:gen-quality-nfe} itself now reports as mean $\pm$ standard deviation, rather than a single seed. The SLERP $<$ TopK/LERP $<$ MDLM ranking on generative perplexity holds at every individual seed and every one of these budgets, with non-overlapping 95\% confidence intervals between all three configurations (full per-seed values and intervals in Appendix~\ref{app:seed_sweep}). MAUVE shows the same directional ranking (SLERP $>$ TopK/LERP $>$ MDLM) at every budget, though with wider and overlapping intervals, consistent with MAUVE's known higher variance at these score magnitudes. $T{=}512$ was not re-run across seeds and is reported from a single seed in Table~\ref{tab:gen-quality-nfe}; we leave a matched 3-seed check at that budget to future work.
\section{Conclusion}
We identified a geometric mismatch at the core of existing soft-masking feedback for MDLMs: the linear (LERP) blend implicitly assumes a Euclidean embedding space, but empirical diagnostics on a released MDLM checkpoint show a near-constant {$\approx$73\textdegree} geodesic angle between the mask direction and top-k predictions alongside frequency-invariant embedding norms, both signatures of a hyperspherical geometry under which LERP embeddings lie off manifold. Spherical Soft-Masking (S-SM) corrects this by aggregating predictions with a Fr\'echet mean and blending via SLERP, restoring the mask token's native norm at every step. Across four inference-time step budgets, SLERP feedback delivers consistently higher MAUVE and lower generative perplexity than both TopK/LERP and the no-feedback baseline, without requiring any change to the underlying training objective, unmasking rule, or confidence schedule. The learned confidence weight itself corroborates the geometric account. The optimizer sustains a substantially larger $\lambda$ under SLERP, indicating the backbone trusts geometrically well-formed feedback more. These results suggest that interpolation geometry, not just the confidence schedule around it, is a first-order design choice for continuous feedback in diffusion language models.
\FloatBarrier
\section{Limitations}

\begin{itemize}
\item \textbf{Scale.} All experiments use a single 169M-parameter backbone. We do not know whether the SLERP-LERP gap, holds at the multi-billion-parameter scale.

\item \textbf{Hyperparameters shared with the baseline.} We inherit the confidence-schedule parameterization, the time-band bounds $(b_l,b_h)$, and the soft-masking activation probability $p_{sm}$ from the LERP baseline without re-tuning them for SLERP; it is possible SLERP has a different optimum for these values that we have not searched.
\item \textbf{Single interpolation family compared.} We compare against the specific LERP/TopK formulation of soft-masking; we do not compare against every alternative continuous-feedback mechanism proposed concurrently, some of which address a different failure mode (uniform corruption, token-wise marginal training) than the one we target (embedding-space geometry).
\item \textbf{Metrics.} MAUVE and generative perplexity are standard but imperfect proxies for generation quality; we do not run human evaluation.
\end{itemize}

\bibliography{custom}

\newpage
\appendix
\clearpage

\appendix

\section{Full Geometric Construction of S-SM}
\label{app:construction}

This appendix gives the complete derivation referenced in Section~5: the
Riemannian log/exp maps used internally, the Karcher-flow update for the
Fr\'echet mean, and the closed-form SLERP blend with its endpoint and norm
properties.

\subsection{Geometric preliminaries}
\label{app:construction:prelim}

For a base point $\boldsymbol{\mu} \in \mathbb{S}^{D-1}$, the Riemannian
logarithm and exponential maps are
\begin{align}
\log_{\boldsymbol{\mu}}(\mathbf{v})
&= \frac{\theta}{\sin\theta}\big(\mathbf{v} - \cos\theta\,\boldsymbol{\mu}\big), \notag \\
&\qquad \theta = \arccos\langle \boldsymbol{\mu}, \mathbf{v}\rangle,
\label{eq:app_logmap} \\
\exp_{\boldsymbol{\mu}}(\mathbf{u})
&= \cos(\|\mathbf{u}\|)\,\boldsymbol{\mu}
+ \sin(\|\mathbf{u}\|)\,\frac{\mathbf{u}}{\|\mathbf{u}\|}, \notag \\
&\qquad \mathbf{u} \in T_{\boldsymbol{\mu}}\mathbb{S}^{D-1}.
\label{eq:app_expmap}
\end{align}
$\log_{\boldsymbol{\mu}}$ flattens a nearby point on the sphere into the
tangent plane at $\boldsymbol{\mu}$; $\exp_{\boldsymbol{\mu}}$ retracts a
tangent vector back onto the sphere along a geodesic. Both maps are used
only as internal building blocks; nothing here is exposed to the rest of the
training pipeline.

\subsection{Weighted Fr\'echet mean via Karcher flow}
\label{app:construction:karcher}

We replace the Euclidean weighted mean $\boldsymbol{\mu}_{\mathrm{LERP}} =
\sum_i \pi_i \mathbf{v}_i$ with the intrinsic spherical (Fr\'echet) mean of the
normalised top-$k$ directions $\hat{\mathbf{v}}_i = \mathbf{v}_i / \|\mathbf{v}_i\|$:
\begin{align}
\boldsymbol{\mu}^{\star}
&= \operatorname*{arg\,min}_{\boldsymbol{\mu} \in \mathbb{S}^{D-1}}
\sum_{i=1}^{k} \pi_i \, d_{\mathrm{g}}(\boldsymbol{\mu}, \hat{\mathbf{v}}_i)^2,
\label{eq:app_frechet} \\
d_{\mathrm{g}}(\mathbf{a}, \mathbf{b}) &= \arccos\langle \mathbf{a}, \mathbf{b}\rangle , \notag
\end{align}
where the sum ranges over the $k$ top-$k$ candidate indices.
Equation~\eqref{eq:app_frechet} has no closed form, so we solve it with
$N_{\mathrm{iter}}$ steps of Karcher fixed-point iteration, initialised at the
top-1 direction $\hat{\mathbf{v}}^{(1)}$:
\begin{equation}
\boldsymbol{\mu} \;\leftarrow\;
\exp_{\boldsymbol{\mu}}\Big(\textstyle\sum_{i=1}^{k} \pi_i \log_{\boldsymbol{\mu}}(\hat{\mathbf{v}}_i)\Big).
\label{eq:app_karcher_update}
\end{equation}
Because $\pi_i$ is a softmax over top-$k$ probabilities it is typically
peaked on the top-1 token, so the iteration converges in very few steps; we
use $N_{\mathrm{iter}} = 3$ by default (ablated in ~\ref{sec:ablations}).
Algorithm~\ref{alg:karcher} spells out Equation~\eqref{eq:app_karcher_update}
step by step, including the numerical guards used in practice.

\begin{algorithm}[t]
\caption{Spherical Soft-Masking Feedback ($\mathrm{sm}^{\text{sph}}_\omega$)}
\label{alg:ssm}
\begin{algorithmic}[1]
\small
\Require
  $E \in \mathbb{R}^{|V| \times D}$ (embedding matrix); \\
  $\mathbf{p} \in \Delta^{|V|-1}$ (predicted token distribution); \\
  mask token index $m$; superposition width $k$; \\
  Karcher iterations $N_{\text{iter}}$; confidence factor $\lambda \in [0,1)$; \\
  numerical tolerances $\epsilon, \delta$
\Ensure
  Soft-masked feedback vector $\mathbf{x}^{\text{sph}} \in \mathbb{R}^{D}$

\State $\mathcal{T} \gets \mathrm{top\text{-}}k(\mathbf{p})$
\State $\pi_i \gets [\mathbf{p}]_i / \sum_{j \in \mathcal{T}} [\mathbf{p}]_j \quad \forall i \in \mathcal{T}$ \Comment{Normalize candidate weights}
\State $\mathbf{m} \gets E[m]$, \quad $r_m \gets \lVert\mathbf{m}\rVert$, \quad $\hat{\mathbf{m}} \gets \mathbf{m} / r_m$
\State $\hat{\mathbf{v}}_i \gets E[i] / \lVert E[i]\rVert \quad \forall i \in \mathcal{T}$ \Comment{Project top-$k$ to $\mathbb{S}^{D-1}$}

\Statex \hrulefill
\Statex \textbf{Phase A: Weighted Fr\'echet Mean via Karcher Flow} (Eq.~\ref{eq:frechet}--\ref{eq:karcher})
\State $\boldsymbol\mu^\star \gets \hat{\mathbf{v}}_{(1)}$ \Comment{Initialize with top-1 prediction}
\For{$\text{step} = 1$ \textbf{to} $N_{\text{iter}}$}
    \For{$i \in \mathcal{T}$}
        \State $c_i \gets \mathrm{clamp}(\langle\boldsymbol\mu^\star, \hat{\mathbf{v}}_i\rangle, -1+\epsilon, 1-\epsilon)$
        \State $\theta_i \gets \arccos c_i$
        \State $\mathbf{u}_i \gets \dfrac{\theta_i}{\sin\theta_i} \big(\hat{\mathbf{v}}_i - c_i \boldsymbol\mu^\star\big)$ \Comment{Log map, Eq.~\ref{eq:logmap}}
    \EndFor
    \State $\bm{\tau} \gets \sum_{i \in \mathcal{T}} \pi_i \mathbf{u}_i$ \Comment{Weighted tangent vector}
    \State $\boldsymbol\mu^\star \gets \cos(\lVert\bm{\tau}\rVert)\boldsymbol\mu^\star + \sin(\lVert\bm{\tau}\rVert) \dfrac{\bm{\tau}}{\lVert\bm{\tau}\rVert}$ \Comment{Exp map, Eq.~\ref{eq:expmap}}
    \State $\boldsymbol\mu^\star \gets \boldsymbol\mu^\star / \lVert\boldsymbol\mu^\star\rVert$ \Comment{Re-normalize against drift}
\EndFor

\Statex \hrulefill
\Statex \textbf{Phase B: Geodesic SLERP Blend} (Eq.~\ref{eq:slerp}--\ref{eq:rescale})
\State $\Omega \gets \arccos\big(\mathrm{clamp}(\langle\hat{\mathbf{m}}, \boldsymbol\mu^\star\rangle, -1+\epsilon, 1-\epsilon)\big)$
\If{$\Omega < \delta$} \Comment{Fallback for near-collinearity}
    \State $\mathbf{s} \gets (1-\lambda)\hat{\mathbf{m}} + \lambda\boldsymbol\mu^\star$
    \State $\mathbf{s} \gets \mathbf{s} / \lVert\mathbf{s}\rVert$
\Else
    \State $\mathbf{s} \gets \dfrac{\sin((1-\lambda)\Omega)}{\sin\Omega}\hat{\mathbf{m}} + \dfrac{\sin(\lambda\Omega)}{\sin\Omega}\boldsymbol\mu^\star$
\EndIf

\Statex \hrulefill
\State $\mathbf{x}^{\text{sph}} \gets r_m \mathbf{s}$ \Comment{Rescale to native token norm}
\Return $\mathbf{x}^{\text{sph}}$
\end{algorithmic}
\end{algorithm}

\begin{algorithm*}[t]
\caption{Weighted Fr\'echet (Karcher) mean on $\mathbb{S}^{D-1}$}
\label{alg:karcher}
\small
\begin{algorithmic}[1]
\Require unit vectors $\{\hat{\mathbf{v}}_i\}_{i=1}^{k} \subset \mathbb{S}^{D-1}$; weights $\boldsymbol{\pi} \in \Delta^{k-1}$; iterations $N_{\mathrm{iter}}$; tolerance $\varepsilon$
\Ensure Fr\'echet mean $\boldsymbol{\mu}^{\star}$
\State $\boldsymbol{\mu} \leftarrow \hat{\mathbf{v}}_1$ \Comment{initialise at top-1 candidate}
\For{$n = 1$ \textbf{to} $N_{\mathrm{iter}}$}
  \For{$i = 1$ \textbf{to} $k$}
    \State $c_i \leftarrow \operatorname{clamp}(\boldsymbol{\mu}^{\top}\hat{\mathbf{v}}_i,\, -1{+}\varepsilon,\, 1{-}\varepsilon)$
    \State $\omega_i \leftarrow \arccos(c_i)$
    \If{$\sin(\omega_i) > \varepsilon$}
      \State $r_i \leftarrow \omega_i / \sin(\omega_i)$
    \Else
      \State $r_i \leftarrow 1$
    \EndIf
    \State $\boldsymbol{\tau}_i \leftarrow r_i \big(\hat{\mathbf{v}}_i - c_i\,\boldsymbol{\mu}\big)$ \Comment{$\log_{\boldsymbol{\mu}}(\hat{\mathbf{v}}_i)$, Eq.~\eqref{eq:app_logmap}}
  \EndFor
  \State $\boldsymbol{\tau} \leftarrow \sum_{i=1}^{k} \pi_i\, \boldsymbol{\tau}_i$ \Comment{weighted tangent vector}
  \State $\boldsymbol{\mu} \leftarrow \cos(\|\boldsymbol{\tau}\|)\,\boldsymbol{\mu} + \sin(\|\boldsymbol{\tau}\|)\, \boldsymbol{\tau}/\|\boldsymbol{\tau}\|$ \Comment{$\exp_{\boldsymbol{\mu}}(\boldsymbol{\tau})$, Eq.~\eqref{eq:app_expmap}}
  \State $\boldsymbol{\mu} \leftarrow \boldsymbol{\mu} / \|\boldsymbol{\mu}\|$ \Comment{re-normalise against drift}
\EndFor
\State \Return $\boldsymbol{\mu}$ as $\boldsymbol{\mu}^{\star}$
\end{algorithmic}
\end{algorithm*}

\subsection{Geodesic blend (SLERP) and norm restoration}
\label{app:construction:slerp}

Let $\hat{\mathbf{m}} = \mathbf{m}/r_m$ be the normalised mask direction and
$\Omega = \arccos\langle \hat{\mathbf{m}}, \boldsymbol{\mu}^{\star}\rangle$ the
geodesic angle between it and the Fr\'echet-mean target. Using the same
confidence weight $\lambda$ as before, but now as an interpolation fraction
along the arc, we compute
\begin{equation}
\mathbf{s} = \frac{\sin\big((1-\lambda)\Omega\big)}{\sin\Omega}\,\hat{\mathbf{m}}
+ \frac{\sin(\lambda\Omega)}{\sin\Omega}\,\boldsymbol{\mu}^{\star},
\label{eq:app_slerp_final}
\end{equation}
and restore the native mask norm before handing the vector to the backbone:
\begin{equation}
\mathbf{x}^{\mathrm{sph}} = r_m\, \mathbf{s}.
\label{eq:app_xsph}
\end{equation}
This construction satisfies properties (P1)-(P3) of Section~4.1 by design:
it reduces to $\mathbf{m}$ at $\lambda=0$ and to $r_m\boldsymbol{\mu}^{\star}$
as $\lambda \to 1$; $\|\mathbf{s}\| = 1$ for all $\lambda$, so the rescaled
output always has norm $r_m$; and $\mathbf{s}$ traces the great-circle arc
between $\hat{\mathbf{m}}$ and $\boldsymbol{\mu}^{\star}$ by construction of
SLERP.

\paragraph{Numerical safeguards.} Inner products used inside $\arccos$ are
clamped to $[-1+\varepsilon,\, 1-\varepsilon]$, and the denominator
$\sin\Omega$ in Equation~\eqref{eq:app_slerp_final} is clamped to
$[\varepsilon, \infty)$. If $\Omega < \delta$ (the mask direction and the
Fr\'echet-mean target are nearly collinear, the regime in which
Equation~\eqref{eq:app_slerp_final} is ill-conditioned), we fall back to a
normalised linear blend,
\begin{equation}
\mathbf{s} = \frac{(1-\lambda)\hat{\mathbf{m}} + \lambda\boldsymbol{\mu}^{\star}}
{\big\|(1-\lambda)\hat{\mathbf{m}} + \lambda\boldsymbol{\mu}^{\star}\big\|},
\label{eq:app_fallback}
\end{equation}
which is safe precisely because the two directions are already almost
identical there, so the chord and the geodesic nearly coincide. We use
$\delta = \varepsilon = 10^{-6}$ in all experiments (Table~\ref{tab:app_hparams}).

\paragraph{Drop-in property.} S-SM changes only the interior of
\textsc{Feedback}: the confidence schedule, the unmasking rule, the training
objective, and the two-pass training procedure (Appendix~\ref{app:procedure:training})
are all unchanged. Every operation in Algorithms~\ref{alg:karcher}
and~\ref{alg:ssm} is differentiable, so gradients flow through S-SM exactly
as they do through $F_{\mathrm{LERP}}$.

\section{Full S-SM Feedback Procedure}
\label{app:procedure}

Sections~5.1 and 5.2 describe S-SM at a single masked position. We now give the
complete batched procedure that replaces $F_{\mathrm{LERP}}$ at every masked
position of a training or sampling batch, and produces a dense embedding
tensor that is fed directly to the backbone. Let $B$ denote the batch size,
so that token IDs form a tensor $\mathbf{x}_t \in \{0,\ldots,V\}^{B\times L}$
and predicted distributions from the no-gradient forward pass form a tensor
$\mathbf{P} \in \mathbb{R}^{B\times L\times V}$, with $\mathbf{p}^{b,l}_{t-1}$
denoting the row at batch index $b$ and position $l$.

\paragraph{Step 1 (identify masked positions).}
Let $\mathcal{M} = \{(b,l) : x_t^{b,l} = [\texttt{MASK}]\}$. For unmasked
positions the output embedding is simply the standard lookup,
$\mathbf{h}^{b,l} = \mathbf{E}[x_t^{b,l}]$.

\paragraph{Step 2 (top-$k$ selection and renormalised weights).}
For each masked position $(b,l) \in \mathcal{M}$, select the top-$k$ indices
of the pass-1 distribution and renormalise:
\begin{align}
(i_1,\ldots,i_k) &= \operatorname{top\text{-}k}\big(\mathbf{p}^{b,l}_{t-1}\big),
\label{eq:app_topk_weights} \\
\pi_j &= \frac{p^{b,l}_{t-1,\,i_j}}{\sum_{j'=1}^{k} p^{b,l}_{t-1,\,i_{j'}}}. \notag
\end{align}

\paragraph{Step 3 (Fr\'echet mean of top-$k$ embeddings).}
Gather and normalise the top-$k$ candidate embeddings,
$\hat{\mathbf{v}}_{i_j} = \mathbf{E}[i_j,:]/\|\mathbf{E}[i_j,:]\|$, and compute
their Fr\'echet mean $\boldsymbol{\mu}^{\star,\,b,l}$ on $\mathbb{S}^{D-1}$
with Algorithm~\ref{alg:karcher}, using weights $\boldsymbol{\pi}$.

\paragraph{Step 4 (SLERP between mask direction and Fr\'echet mean).}
With $\hat{\mathbf{m}} = \mathbf{m}/r_m$ as in Appendix~\ref{app:construction:slerp},
compute
\begin{equation}
\mathbf{s}^{b,l} = \operatorname{SLERP}\big(\hat{\mathbf{m}},\, \boldsymbol{\mu}^{\star,\,b,l},\, \lambda^{b,l}\big)
\label{eq:app_slerp_bl}
\end{equation}
via Equation~\eqref{eq:app_slerp_final} (falling back to
Equation~\eqref{eq:app_fallback} when $\Omega^{b,l} < \delta$), where
$\lambda^{b,l}$ is the position-specific confidence weight of
Appendix~\ref{app:procedure:lambda}.

\paragraph{Step 5 (norm restoration).}
Rescale by the mask token's native norm, exactly as in Equation~\eqref{eq:app_xsph}:
$\mathbf{h}^{b,l} = r_m\, \mathbf{s}^{b,l}$.

\paragraph{Step 6 (assembly).}
The tensor handed to the backbone is
\begin{equation}
\mathbf{X}[b,l,:] =
\begin{cases}
\mathbf{h}^{b,l}, & (b,l) \in \mathcal{M}, \\[2pt]
\mathbf{E}[x_t^{b,l}], & \text{otherwise},
\end{cases}
\label{eq:app_assembly}
\end{equation}
with $\mathbf{X} \in \mathbb{R}^{B\times L\times D}$, matching the shape of a
standard embedding lookup. Algorithm~\ref{alg:ssm-batched}  summarises Steps 1-6.
\begin{algorithm*}[t]
\caption{S-SM Feedback (batched)}
\label{alg:ssm-batched}
\begin{algorithmic}[1]
\small
\Require token IDs $\mathbf{x}_t \in \{0,\ldots,V\}^{B\times L}$; pass-1 distributions $\mathbf{P} \in \mathbb{R}^{B\times L\times V}$; embedding table $\mathbf{E} \in \mathbb{R}^{V\times D}$; mask token id; top-$k$; Karcher iterations $N_{\mathrm{iter}}$; tolerances $\varepsilon, \delta$
\Ensure embedding tensor $\mathbf{X} \in \mathbb{R}^{B\times L\times D}$
\State $\mathcal{M} \gets \{(b,l) : x_t^{b,l} = [\texttt{MASK}]\}$
\State $\mathbf{X} \gets \mathbf{E}[\mathbf{x}_t]$ \Comment{standard lookup for all positions}
\State Compute $\lambda^{b,l}$ for all $(b,l)\in\mathcal{M}$ via Eq.~\eqref{eq:app_lambda_bl}
\For{\textbf{each} $(b,l) \in \mathcal{M}$}
  \State $(i_1,\ldots,i_k),\, \boldsymbol{\pi} \gets$ top-$k$ \& renormalise $\mathbf{p}^{b,l}_{t-1}$ \Comment{Eq.~\eqref{eq:app_topk_weights}}
  \State $\hat{\mathbf{v}}_{i_j} \gets \mathbf{E}[i_j,:]/\|\mathbf{E}[i_j,:]\|$ for $j=1,\ldots,k$
  \State $\boldsymbol{\mu}^{\star} \gets \textsc{KarcherMean}(\{\hat{\mathbf{v}}_{i_j}\}, \boldsymbol{\pi}, N_{\mathrm{iter}}, \varepsilon)$ \Comment{Alg.~\ref{alg:karcher}}
  \State $\hat{\mathbf{m}} \gets \mathbf{m}/r_m$
  \State $\Omega \gets \arccos\big(\operatorname{clamp}(\hat{\mathbf{m}}^{\top}\boldsymbol{\mu}^{\star}, -1{+}\varepsilon, 1{-}\varepsilon)\big)$
  \If{$\Omega < \delta$}
    \State $\mathbf{s} \gets \big[(1-\lambda^{b,l})\hat{\mathbf{m}} + \lambda^{b,l}\boldsymbol{\mu}^{\star}\big] / \|\cdot\|$ \Comment{Eq.~\eqref{eq:app_fallback}}
  \Else
    \State $\mathbf{s} \gets \operatorname{SLERP}(\hat{\mathbf{m}}, \boldsymbol{\mu}^{\star}, \lambda^{b,l})$ \Comment{Eq.~\eqref{eq:app_slerp_final}}
  \EndIf
  \State $\mathbf{X}[b,l,:] \gets r_m \cdot \mathbf{s}$ \Comment{norm restoration, Eq.~\eqref{eq:app_xsph}}
\EndFor
\State \Return $\mathbf{X}$
\end{algorithmic}
\end{algorithm*}

\subsection{Confidence-based interpolation weight}
\label{app:procedure:lambda}

The confidence weight $\lambda^{b,l}$ used in Equation~\eqref{eq:app_slerp_bl}
follows the same schedule as the underlying soft-masking formulation
(Section~3): for each masked position, the entropy of the pass-1 distribution
is computed,
\begin{equation}
H^{b,l} = -\sum_{v=1}^{V} p^{b,l}_{t-1,v} \log p^{b,l}_{t-1,v},
\label{eq:app_entropy}
\end{equation}
and the mixing weight is
\begin{equation}
\lambda^{b,l} = \omega_s \cdot \sigma\big(\omega_a\,(-H^{b,l} - \omega_b)\big),
\label{eq:app_lambda_bl}
\end{equation}
with $\sigma(\cdot)$ the sigmoid function. The three scalars $\omega_s$
(scale), $\omega_a$ (steepness), and $\omega_b$ (centre) are shared across all
positions and timesteps and are trained jointly with the backbone, exactly as
in Section~3. In our implementation $\omega_s$ is parameterised through a
logit so that $\omega_s \in (0,1)$, while $\omega_a$ and $\omega_b$ are
parameterised via the softplus function to enforce positivity of the
steepness and of the (shifted) centre. Default initialisations are
$\omega_s^{(0)} = 0.5$, $\omega_b^{(0)} = -4.0$, and $\omega_a^{(0)} =
10/1.5 \approx 6.67$. An optional \texttt{fixed\_lambda} override replaces
the learned schedule of Equation~\eqref{eq:app_lambda_bl} with a constant
$\lambda$.

\subsection{Integration into the two-pass training procedure}
\label{app:procedure:training}

Training follows the two-pass procedure of Section~6.1; our modification
changes only the representation fed to the second pass.

\paragraph{Pass 1 (no gradient).} Given corrupted tokens $\mathbf{x}_t$, the
backbone produces $\mathbf{P} = g_\theta(\mathbf{x}_t)$ under
\texttt{torch.no\_grad()}, avoiding construction of the autograd graph for a
forward pass whose output is used only as data.

\paragraph{Pass 2 (gradient).} Algorithm~\ref{alg:ssm} maps
$(\mathbf{x}_t, \mathbf{P})$ to an embedding tensor $\mathbf{X} \in
\mathbb{R}^{B\times L\times D}$, which is passed directly to the backbone,
$\hat{\mathbf{p}}^{1:L} = g_\theta(\mathbf{X})$, and the standard MDLM loss
$\mathcal{L}(\theta,\omega)$ is backpropagated to update both the backbone
parameters $\theta$ and the confidence-schedule parameters
$\omega = (\omega_s,\omega_a,\omega_b)$, with separate learning rates
$\eta_{\mathrm{bb}}$ and $\eta_{\mathrm{sm}}$ (Table~\ref{tab:app_hparams}).

\paragraph{Stochastic gating and time-band restriction.} As in Section~6.1,
the two-pass soft-masking path is activated with probability
$p_{\mathrm{sm}}$ (default $0.5$) and only when the batch-mean timestep
$\bar t = \frac{1}{B}\sum_b t_b$ falls within $[b_l, b_h]$ (default
$[0.2, 0.8]$). When the gate is off or the batch is out of band, a standard
single-pass forward without feedback is used. The gating decision is
deterministic given the global training step (seeded by the step index) so
that all ranks take identical branching decisions in distributed
data-parallel training (Appendix~\ref{app:distributed}).

\subsection{Integration into sampling}
\label{app:procedure:sampling}

During inference with the DDPM caching sampler, S-SM feedback
(Algorithm~\ref{alg:ssm}) is applied at every denoising step whose timestep
$t$ lies within the band $[b_l, b_h]$. Outside the band, including the final
noise-removal step at $t \approx \varepsilon$, a standard forward pass
without feedback is used. The cached log-probability tensor from the
previous denoising step serves as $\mathbf{P}$ in Step~2.

\section{Backbone Embedding Layer Modification}
\label{app:backbone}

The S-SM feedback pathway produces a pre-embedded tensor $\mathbf{X} \in
\mathbb{R}^{B\times L\times D}$ rather than token IDs or a probability
distribution over the vocabulary. We extend the backbone's embedding layer
with a pass-through branch that detects this case (a 3-D input whose last
dimension equals $D$) and returns the input unchanged, so that no
modification to the Transformer blocks themselves is required. The dispatch
logic is:
\begin{enumerate}
\item \textbf{Token IDs} ($\mathbf{x} \in \{0,\ldots,V\}^{B\times L}$):
standard lookup $\mathbf{E}[\mathbf{x}]$.
\item \textbf{Sparse top-$k$ representation} (indices and weights): weighted
gather-and-sum, used by $F_{\mathrm{LERP}}$.
\item \textbf{Pre-embedded tensor} ($\mathbf{X} \in \mathbb{R}^{B\times L
\times D}$): pass-through, used by S-SM (Algorithm~\ref{alg:ssm}).
\item \textbf{Dense probability distribution} ($\mathbf{Q} \in
\mathbb{R}^{B\times L\times V}$): matrix multiply $\mathbf{Q}\mathbf{E}$,
used by the original (dense, non-top-$k$) simplex-space soft-masking
implementation of \citet{hersche2026soft}; retained for backward
compatibility but not used by any experiment reported in this paper.
\end{enumerate}

\section{Comparison with Linear Soft-Masking}
\label{app:comparison}

Table~\ref{tab:app_slerp_vs_lerp} summarises the key differences between the
Euclidean baseline ($\mathrm{TopK}/\mathrm{LERP}$, Section~3) and S-SM.

\begin{table}[t]
\centering
\small
\begin{tabularx}{\linewidth}{@{}l X X@{}}
\toprule
 & \textbf{TopK/LERP} & \textbf{S-SM (ours)} \\
\midrule
Interpolation domain & Probability simplex $\Delta^{|\mathcal V|-1}$ &
Unit hypersphere $\mathbb{S}^{D-1}$ \\
Interpolation type & Linear (chord) & Spherical (geodesic, SLERP) \\
Multi-token aggregation & Weighted Euclidean mean $\boldsymbol{\mu}_{\mathrm{LERP}}$
& Fr\'echet mean $\boldsymbol{\mu}^{\star}$ on $\mathbb{S}^{D-1}$ \\
Backbone input & Sparse $(B,L,k{+}1)$ indices + weights &
Dense $(B,L,D)$ embeddings \\
Additional hyperparameters & None & $N_{\mathrm{iter}}$ (Karcher iterations) \\
Compute overhead & Negligible & $k \cdot N_{\mathrm{iter}}$ trigonometric
ops per masked position \\
\bottomrule
\end{tabularx}
\caption{Comparison of linear (TopK/LERP) and spherical (S-SM) soft-masking.}
\label{tab:app_slerp_vs_lerp}
\end{table}

\section{Implementation Details}
\label{sec:impl_details}

\subsection{Hyperparameters}
\label{app:hparams}

Table~\ref{tab:app_hparams} lists the S-SM hyperparameters and their default
values used throughout Section~6.

\begin{table}[t]
\centering
\small
\begin{tabularx}{\linewidth}{@{}l X l@{}}
\toprule
\textbf{Hyperparameter} & \textbf{Description} & \textbf{Default} \\
\midrule
\texttt{feedback\_alg} & Feedback operator selector & \texttt{s\_sm} \\
$k$ & Top-$k$ candidates & 3 \\
$N_{\mathrm{iter}}$ & Karcher iterations & 3 \\
$\omega_s^{(0)}$ & Initial scale & 0.5 \\
$\omega_b^{(0)}$ & Initial sigmoid centre & $-4.0$ \\
$\omega_a^{(0)}$ & Initial sigmoid steepness & 6.67 \\
$p_{\mathrm{sm}}$ & Soft-masking activation probability & 0.5 \\
$b_l / b_h$ & Timestep band & 0.2 / 0.8 \\
$\eta_{\mathrm{sm}}$ & Confidence-parameter learning rate & $10^{-2}$ \\
$\eta_{\mathrm{bb}}$ & Backbone learning rate & $3\times10^{-5}$ \\
$\varepsilon$ & Numerical clamping tolerance & $10^{-6}$ \\
$\delta$ & SLERP near-collinearity threshold & $10^{-6}$ \\
\bottomrule
\end{tabularx}
\caption{S-SM hyperparameters and default values.}
\label{tab:app_hparams}
\end{table}

\subsection{MAUVE Implementation Details}
\label{sec:mauve_details}

All MAUVE and generative-perplexity results (Table~\ref{tab:gen-quality-nfe},
Table~\ref{tab:app_seed_ci}) are computed from generations produced by the
final checkpoint of each 7{,}000-step continued pre-training run
(one checkpoint per seed, per feedback configuration), not from checkpoints
sampled across training.

We compute MAUVE using the standard \texttt{mauve-text} package with its default settings, using the library's default GPT-2 featurization model (note: this is separate from the larger GPT-2 model we use for perplexity scoring) and default clustering hyperparameters. For reference text, we sample from the OpenWebText validation split, matching the count to our generated samples (5,000 per configuration) in all reported results.

We don't pass an explicit seed into MAUVE itself. Reproducibility is instead handled globally at the start of each run, which means MAUVE's internal k-means clustering isn't independently seeded across our three evaluation seeds. We suspect this is part of why MAUVE shows higher seed-to-seed variance than Gen PPL (see Section~\ref{sec:seed_robustness}).

\section{Numerical Stability}
\label{app:numerical}

All trigonometric computations ($\arccos$, $\sin$) within Algorithms
\ref{alg:karcher} and~\ref{alg:ssm} are performed in \texttt{float32}
precision, even when the backbone otherwise operates in \texttt{bfloat16}.
The following guards ensure numerical stability:
\begin{itemize}
\item Every cosine similarity used inside $\arccos$ is clamped to
$[-1+\varepsilon,\, 1-\varepsilon]$ before the inverse cosine is applied,
preventing NaN gradients at the boundary $|\cos| = 1$.
\item The denominator $\sin\Omega$ in the SLERP formula
(Equation~\eqref{eq:app_slerp_final}) is clamped to $[\varepsilon, \infty)$.
\item When $\Omega < \delta$ (the mask direction and the Fr\'echet mean are
nearly identical), S-SM falls back to the normalised linear blend of
Equation~\eqref{eq:app_fallback}, avoiding division by a near-zero
$\sin\Omega$.
\item In the Karcher-mean log map (Algorithm~\ref{alg:karcher}), the
ratio $\omega_i / \sin(\omega_i)$ is replaced by $1$ whenever
$\sin(\omega_i) < \varepsilon$, which is the correct limit as
$\omega_i \to 0$.
\end{itemize}

\section{Computational Efficiency}
\label{app:efficiency}

Three optimisations reduce the overhead of the S-SM pathway relative to a
na\"ive implementation:
\begin{enumerate}
\item \textbf{Masked-only computation.} All operations in
Algorithm~\ref{alg:ssm} (top-$k$ selection, embedding gather, Fr\'echet mean,
SLERP) are applied only to the $M = |\mathcal{M}|$ masked positions rather
than the full $B\times L$ tensor. At a typical masking rate of 30-70\%,
this reduces FLOPs and intermediate memory by a factor proportional to the
masking fraction.
\item \textbf{No-gradient first pass.} The feedback forward pass (Pass~1,
Appendix~\ref{app:procedure:training}) is wrapped in
\texttt{torch.no\_grad()}, which avoids constructing the backward graph and
substantially reduces peak activation memory.
\item \textbf{Selective precision.} Only the trigonometric core
($\arccos$, $\sin$, normalisation) runs in \texttt{float32}; the embedding
gathers and the final norm rescaling (Equation~\eqref{eq:app_xsph}) operate
in the backbone's native \texttt{bfloat16}.
\end{enumerate}

\section{Distributed Training Considerations}
\label{app:distributed}

The S-SM pathway is fully compatible with PyTorch DDP (Distributed Data
Parallel). All operations in Algorithms~\ref{alg:karcher} and~\ref{alg:ssm}
are per-sample and elementwise, with no cross-rank communication inside the
feedback computation. To ensure that all ranks take the same soft-masking
branching decision at every step (required for correct gradient
all-reduction), the stochastic gate of Appendix~\ref{app:procedure:training}
is seeded by the global training step rather than by a per-rank random draw,
and the time-band test uses the all-reduced global mean of $t$ across ranks.
The confidence-schedule parameters $\omega = (\omega_s,\omega_a,\omega_b)$
receive gradients on every soft-masking step and are synchronised through the
standard DDP gradient all-reduce; because soft-masking is gated
stochastically, \texttt{find\_unused\_parameters=True} is required. On steps
where the soft-masking branch is not taken, $\omega$ receives no gradient. We run all our experiments on a GPU cluster with four NVIDIA A6000s. 

\section{Full 3-Seed Sweep: Per-Seed Results and Confidence Intervals}
\label{app:seed_sweep}

To more thoroughly assess seed robustness we repeat generation under three training/generation seeds for all three feedback configurations (MDLM baseline, TopK/LERP, SLERP) at $T\in\{64,128,256\}$. Table~\ref{tab:app_seed_raw} reports every individual run; Table~\ref{tab:app_seed_ci} aggregates these into per-configuration means with 95\% confidence intervals (Student's $t$, $n=3$, computed from the seed-to-seed sample standard deviation). 

In Table \ref{tab:app_seed_ci}, on Gen PPL, the SLERP / TopK/LERP / MDLM intervals are mutually non-overlapping at every budget. On MAUVE the same ranking holds directionally at every budget (SLERP $>$ TopK/LERP $>$ MDLM in mean), but intervals are wider and frequently overlapping between adjacent configurations (most consistently between TopK/LERP and SLERP), reflecting MAUVE's known higher variance at these score magnitudes.

\begin{table}[t]
\centering
\scriptsize
\setlength{\tabcolsep}{3pt}
\begin{tabular}{llrrrr}
\toprule
\textbf{Arm} & \textbf{Seed} & \textbf{NFE} & \textbf{Gen PPL} & \textbf{Entropy} & \textbf{MAUVE} \\
\midrule
MDLM & 1 & 1/16& 86.5890 & 5.5833 & 0.008825 \\
MDLM & 2 & 1/16& 88.1641 & 5.5872 & 0.008535 \\
MDLM & 3 & 1/16& 86.9929 & 5.5841 & 0.008415 \\
MDLM & 1 & 1/8& 73.5284 & 5.5543 & 0.011559 \\
MDLM & 2 & 1/8& 73.1973 & 5.5526 & 0.011425 \\
MDLM & 3 & 1/8& 72.7227 & 5.5515 & 0.011944 \\
MDLM & 1 & 1/4& 66.0873 & 5.5272 & 0.021002 \\
MDLM & 2 & 1/4& 66.2462 & 5.5271 & 0.015042 \\
MDLM & 3 & 1/4& 66.2172 & 5.5256 & 0.016703 \\
\addlinespace
TopK/LERP & 1 & 1/16& 82.8026 & 5.5846 & 0.009850 \\
TopK/LERP & 2 & 1/16& 82.8806 & 5.5846 & 0.009039 \\
TopK/LERP & 3 & 1/16& 82.0693 & 5.5819 & 0.008481 \\
TopK/LERP & 1 & 1/8& 69.1745 & 5.5504 & 0.014883 \\
TopK/LERP & 2 & 1/8& 69.0338 & 5.5517 & 0.014448 \\
TopK/LERP & 3 & 1/8& 68.9927 & 5.5516 & 0.015215 \\
TopK/LERP & 1 & 1/4& 62.4244 & 5.5256 & 0.025190 \\
TopK/LERP & 2 & 1/4& 62.8213 & 5.5234 & 0.023010 \\
TopK/LERP & 3 & 1/4& 62.1231 & 5.5278 & 0.021034 \\
\addlinespace
SLERP & 1 & 1/16& 73.2043 & 5.5682 & 0.012243 \\
SLERP & 2 & 1/16& 72.0246 & 5.5650 & 0.012137 \\
SLERP & 3 & 1/16& 72.2609 & 5.5643 & 0.010738 \\
SLERP & 1 & 1/8& 60.0222 & 5.5310 & 0.016672 \\
SLERP & 2 & 1/8& 59.3822 & 5.5289 & 0.020155 \\
SLERP & 3 & 1/8& 59.8854 & 5.5288 & 0.020005 \\
SLERP & 1 & 1/4& 54.2024 & 5.5082 & 0.040841 \\
SLERP & 2 & 1/4& 53.1495 & 5.5061 & 0.035710 \\
SLERP & 3 & 1/4& 53.8278 & 5.5040 & 0.029775 \\
\bottomrule
\end{tabular}
\caption{Individual per-seed results for all three feedback configurations at $T\in\{64,128,256\}$.}
\label{tab:app_seed_raw}
\end{table}

\begin{table}[t]
\centering
\scriptsize
\setlength{\tabcolsep}{3pt}
\begin{tabular}{llrrr}
\toprule
\textbf{Arm} & \textbf{NFE} & \textbf{Gen PPL} $\downarrow$ & \textbf{Entropy} & \textbf{MAUVE} $\uparrow$ \\
\midrule
MDLM & 1/16& $87.25 \pm 2.03$ & $5.5849 \pm 0.0051$ & $0.00859 \pm 0.00052$ \\
MDLM & 1/8& $73.15 \pm 1.01$ & $5.5528 \pm 0.0035$ & $0.01164 \pm 0.00067$ \\
MDLM & 1/4& $66.18 \pm 0.21$ & $5.5266 \pm 0.0022$ & $0.01758 \pm 0.00764$ \\
\addlinespace
TopK/LERP & 1/16& $82.58 \pm 1.11$ & $5.5837 \pm 0.0039$ & $0.00912 \pm 0.00171$ \\
TopK/LERP & 1/8& $69.07 \pm 0.24$ & $5.5512 \pm 0.0018$ & $0.01485 \pm 0.00096$\\
TopK/LERP & 1/4& $62.46 \pm 0.87$ & $5.5256 \pm 0.0055$ & $0.02308 \pm 0.00516$ \\
\addlinespace
SLERP & 1/16& $72.50 \pm 1.55$ & $5.5658 \pm 0.0052$ & $0.01171 \pm 0.00209$ \\
SLERP & 1/8& $59.76 \pm 0.84$ & $5.5296 \pm 0.0031$ & $0.01894 \pm 0.00489$ \\
SLERP & 1/4& $53.73 \pm 1.33$ & $5.5061 \pm 0.0052$ & $0.03544 \pm 0.01376$ \\
\bottomrule
\end{tabular}
\caption{Mean $\pm$ 95\% CI (Student's $t$, $n{=}3$, sample standard deviation across seeds) for each configuration and NFE budget.}
\label{tab:app_seed_ci}
\end{table}


\subsection{Computational Overhead}
\label{sec:overhead}

At one masked position, $\mathcal{F}_{\text{LERP}}$ forms
$\mu_{\text{LERP}} = \sum_i \pi_i v_i$ and blends it with $\mathbf{m}$,
costing $(2k+1)D$ flops and no transcendental calls. S-SM adds two
phases on top of this. Each Karcher iteration (Alg.~2) does, per
candidate, one dot product, one \texttt{arccos}, one \texttt{sin}, and
one log-map combination, plus a single exp-map retraction, for
$\approx 6kD$ flops and $2k+2$ trigonometric calls. The SLERP blend
(Eq.~7) then costs $O(D)$ flops and a further 4 calls (one
\texttt{arccos}, three \texttt{sin}), independent of $k$ and
$N_{\text{iter}}$. Totalling,
\[
\text{FLOPs}_{\text{S-SM}} \approx 6\,N_{\text{iter}}kD,
\]
\[
\qquad
\text{trig} = N_{\text{iter}}(2k+2) + 4 .
\]

\end{document}